\documentclass[conference]{IEEEtran}
\usepackage{cite}
\usepackage{amsmath,amssymb,amsfonts}
\usepackage{algorithmic}
\usepackage{graphicx}
\usepackage{textcomp}
\usepackage{xcolor}
\usepackage{multirow}
\usepackage{subcaption}

\usepackage[draft=false]{defs-common}
\usepackage{defs-custom}

\newcommand{\sectionpre}[0]{\vspace{-4pt}}
\newcommand{\sectionpost}[0]{\vspace{-3pt}}
\newcommand{\subsectionpre}[0]{\vspace{-3pt}}
\newcommand{\subsectionpost}[0]{\vspace{-3pt}}

\title{Dungeon and Platformer Level Blending and Generation using Conditional VAEs}

\author{\IEEEauthorblockN{Anurag Sarkar}
\IEEEauthorblockA{\textit{Northeastern University} \\
Boston, MA, USA \\
sarkar.an@northeastern.edu}
\and
\IEEEauthorblockN{Seth Cooper}
\IEEEauthorblockA{\textit{Northeastern University} \\
Boston, MA, USA \\
se.cooper@northeastern.edu}
}

\begin{document}

\twocolumn[{%
\renewcommand\twocolumn[1][]{#1}%
\maketitle
\begin{center}
    \centering
    \captionsetup{type=figure}
    \includegraphics[width=\textwidth,height=11.25cm]
    {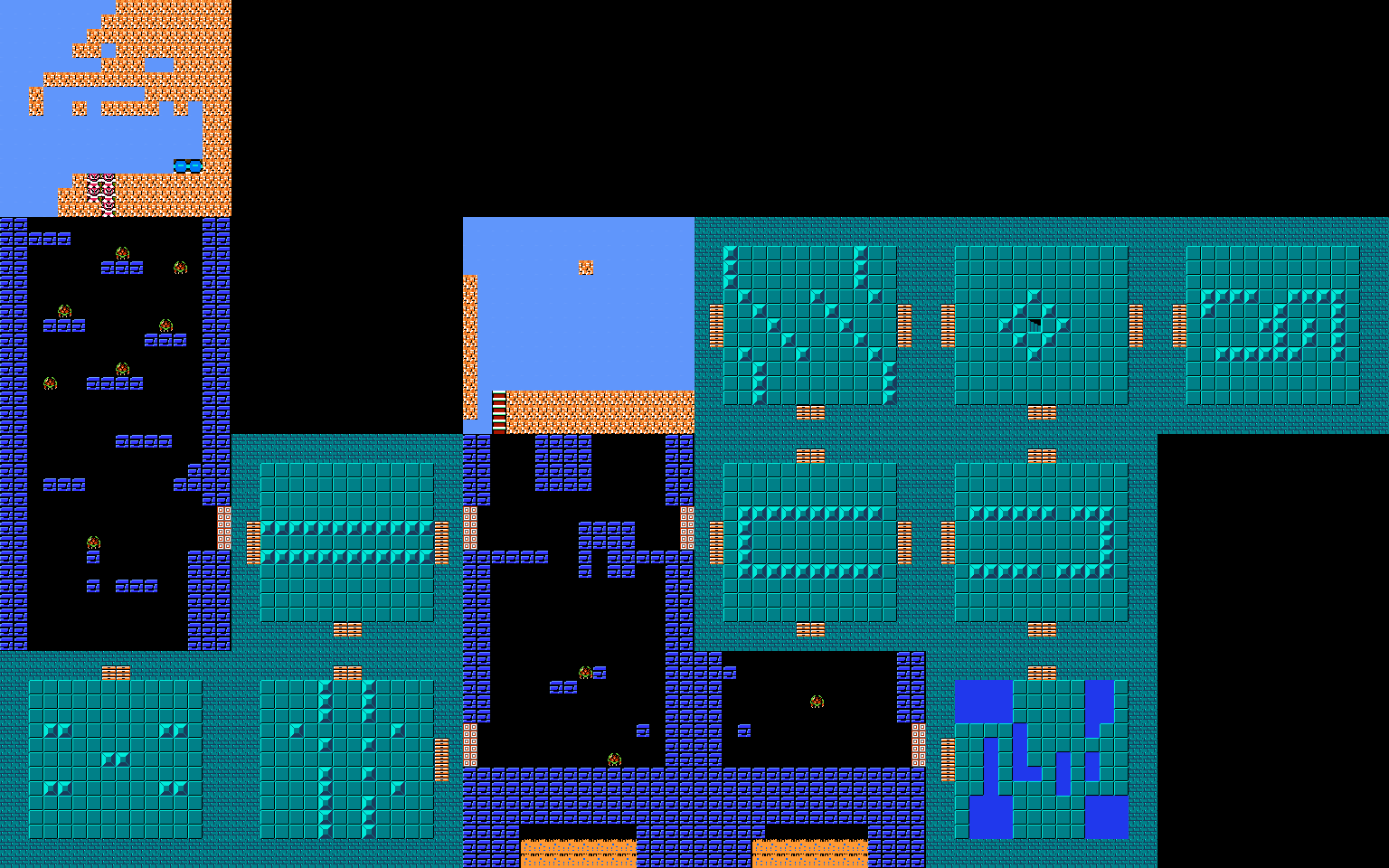}
    \captionof{figure}{Example blended Zelda--Metroid--Mega Man level}
\end{center}%
}]

\begin{abstract}
Variational autoencoders (VAEs) have been used in prior works for generating and blending levels from different games. To add controllability to these models, conditional VAEs (CVAEs) were recently shown capable of generating output that can be modified using labels specifying desired content, albeit working with segments of levels and platformers exclusively. We expand these works by using CVAEs for generating whole platformer and dungeon levels, and blending levels across these genres. We show that CVAEs can reliably control door placement in dungeons and progression direction in platformer levels. Thus, by using appropriate labels, our approach can generate whole dungeons and platformer levels of interconnected rooms and segments respectively as well as levels that blend dungeons and platformers. We demonstrate our approach using \textit{The Legend of Zelda, Metroid, Mega Man} and \textit{Lode Runner}.
\end{abstract}


\newcommand{\XFIGUREwelcome}{
\begin{figure}[]
\centering
\includegraphics[width=1\columnwidth]{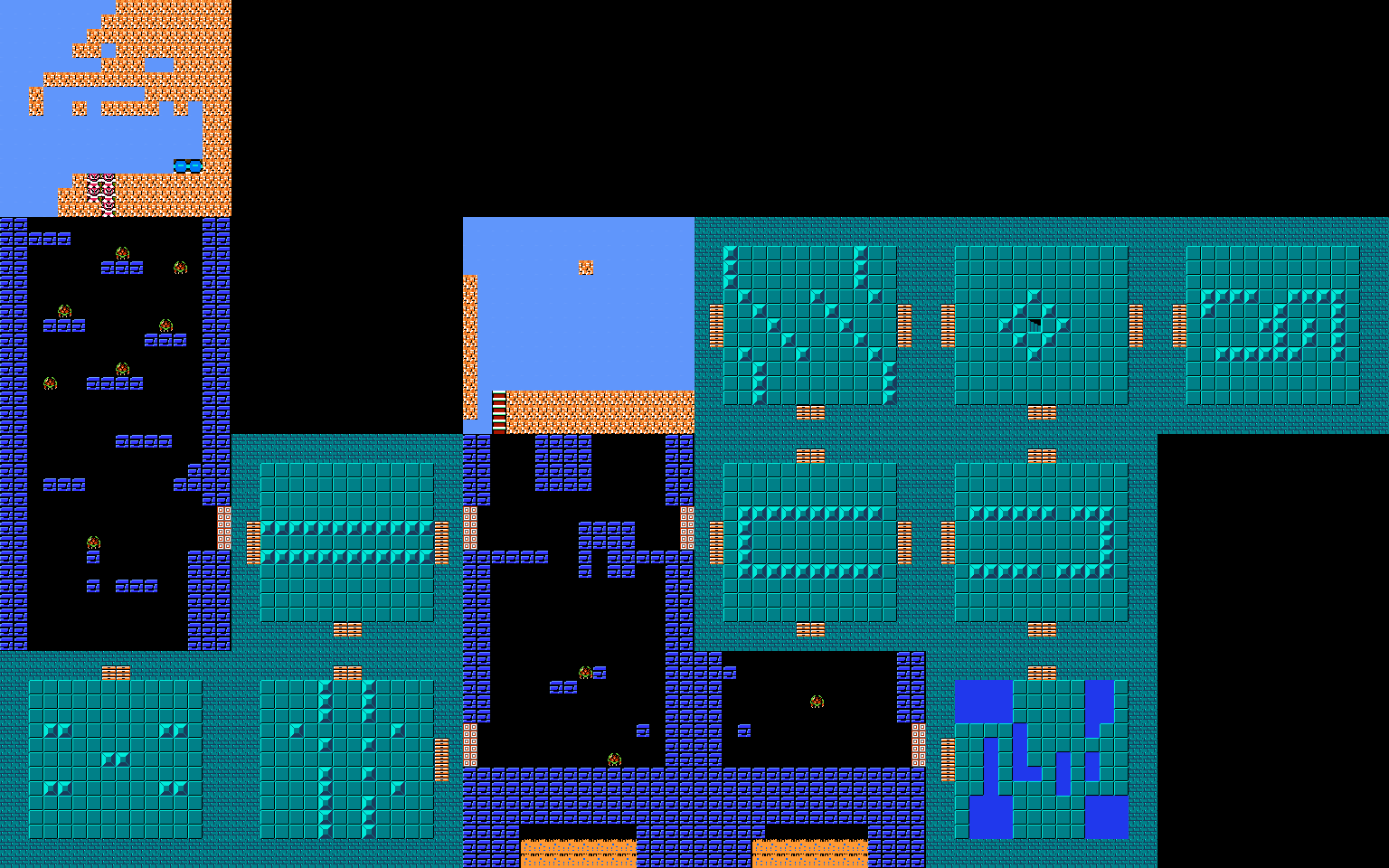}
\caption{\label{XFIGUREwelcome} Example blend of Zelda--Metroid--Mega Man.}
\end{figure}
}

\newcommand{\XFIGUREzelda}{
\begin{figure}[t]
\centering
\includegraphics[width=1\columnwidth]{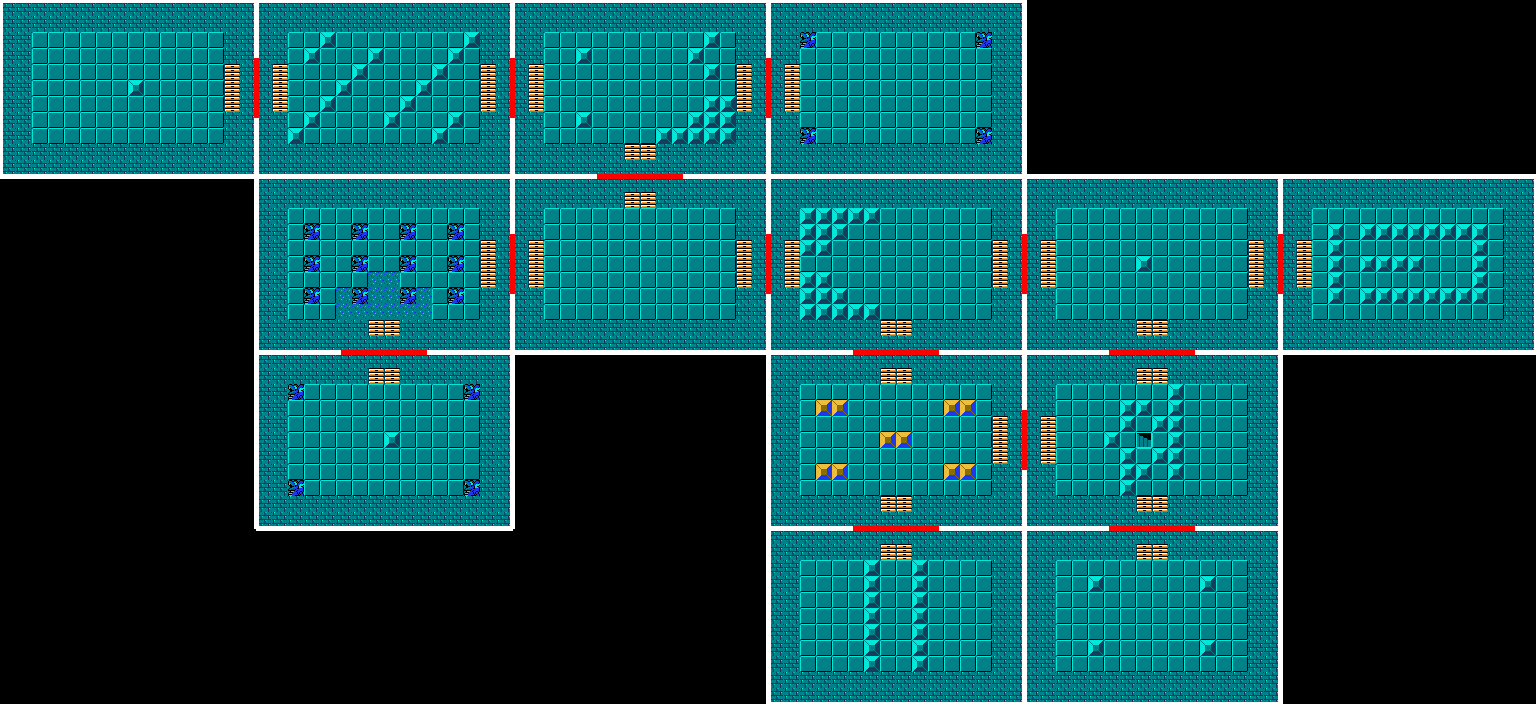}
\caption{\label{XFIGUREzelda} Example Zelda level.}
\end{figure}
}

\newcommand{\XFIGUREzeldalode}{
\begin{figure}[t]
\centering
\includegraphics[width=1\columnwidth]{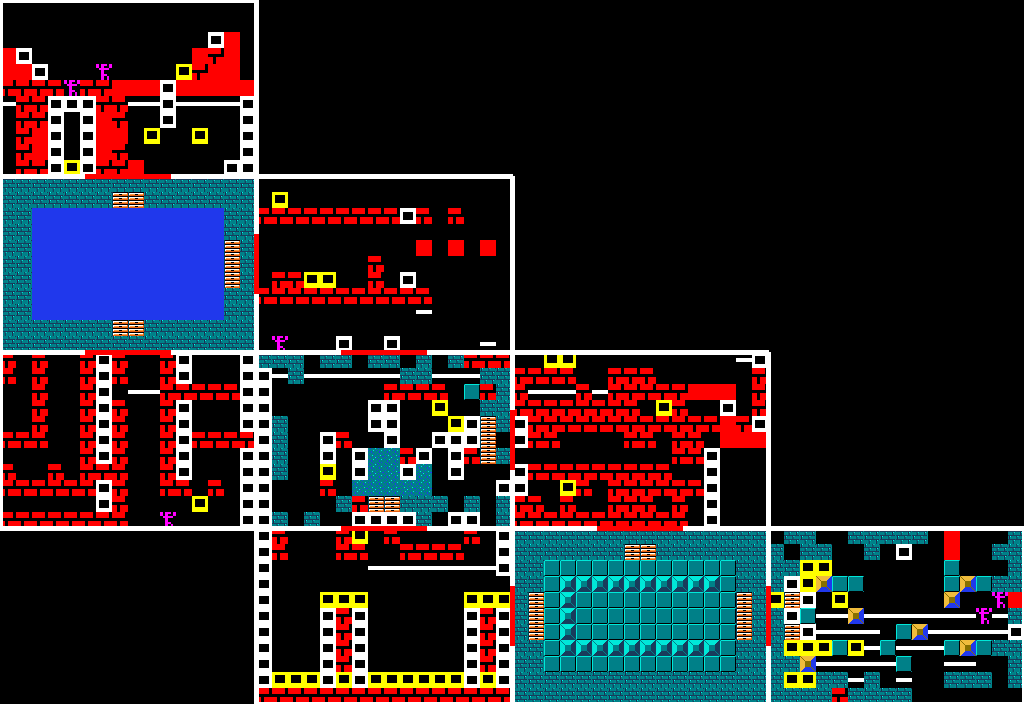}
\caption{\label{XFIGUREzeldalode} Example blended Zelda--Lode Runner level.}
\end{figure}
}

\newcommand{\XFIGUREmetroid}{
\begin{figure}[t]
\centering
\includegraphics[width=1\columnwidth]{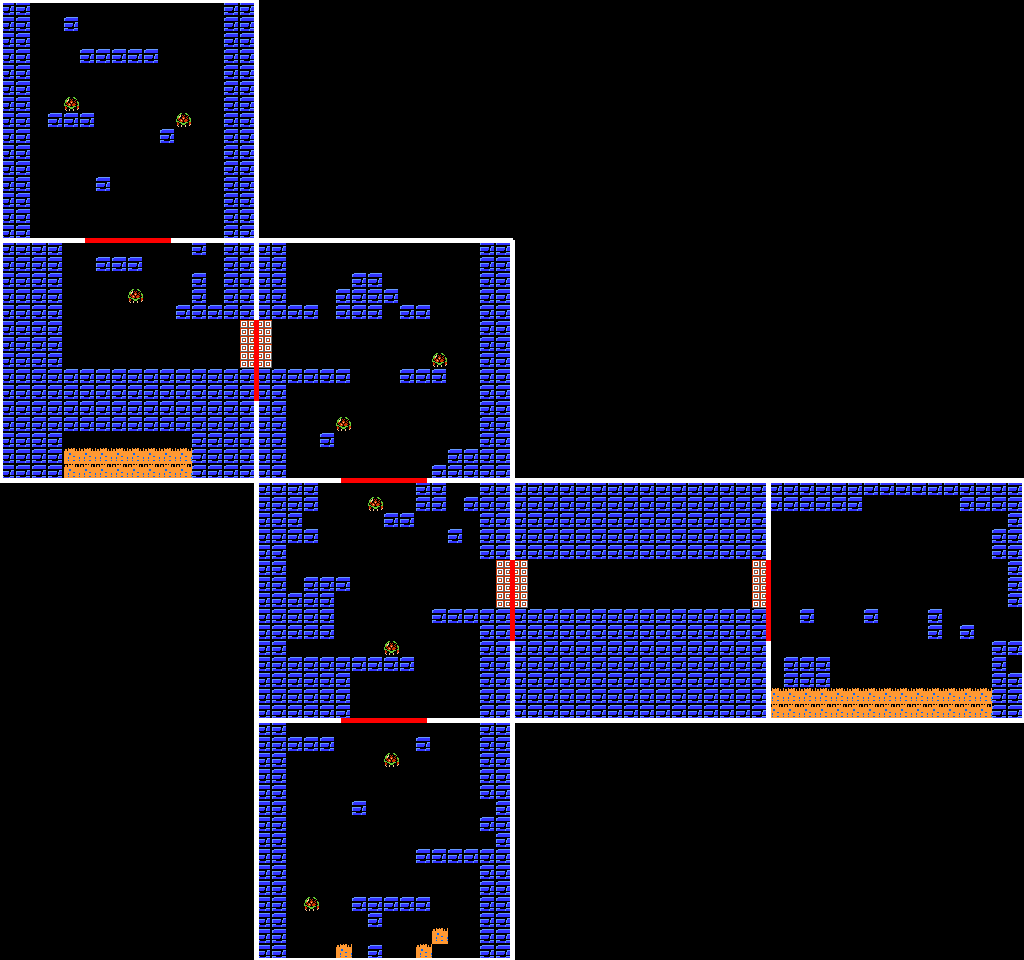}
\caption{\label{XFIGUREmetroid} Example Metroid level.}
\end{figure}
}

\newcommand{\XFIGUREzeldamet}{
\begin{figure}[t]
\centering
\includegraphics[width=1\columnwidth]{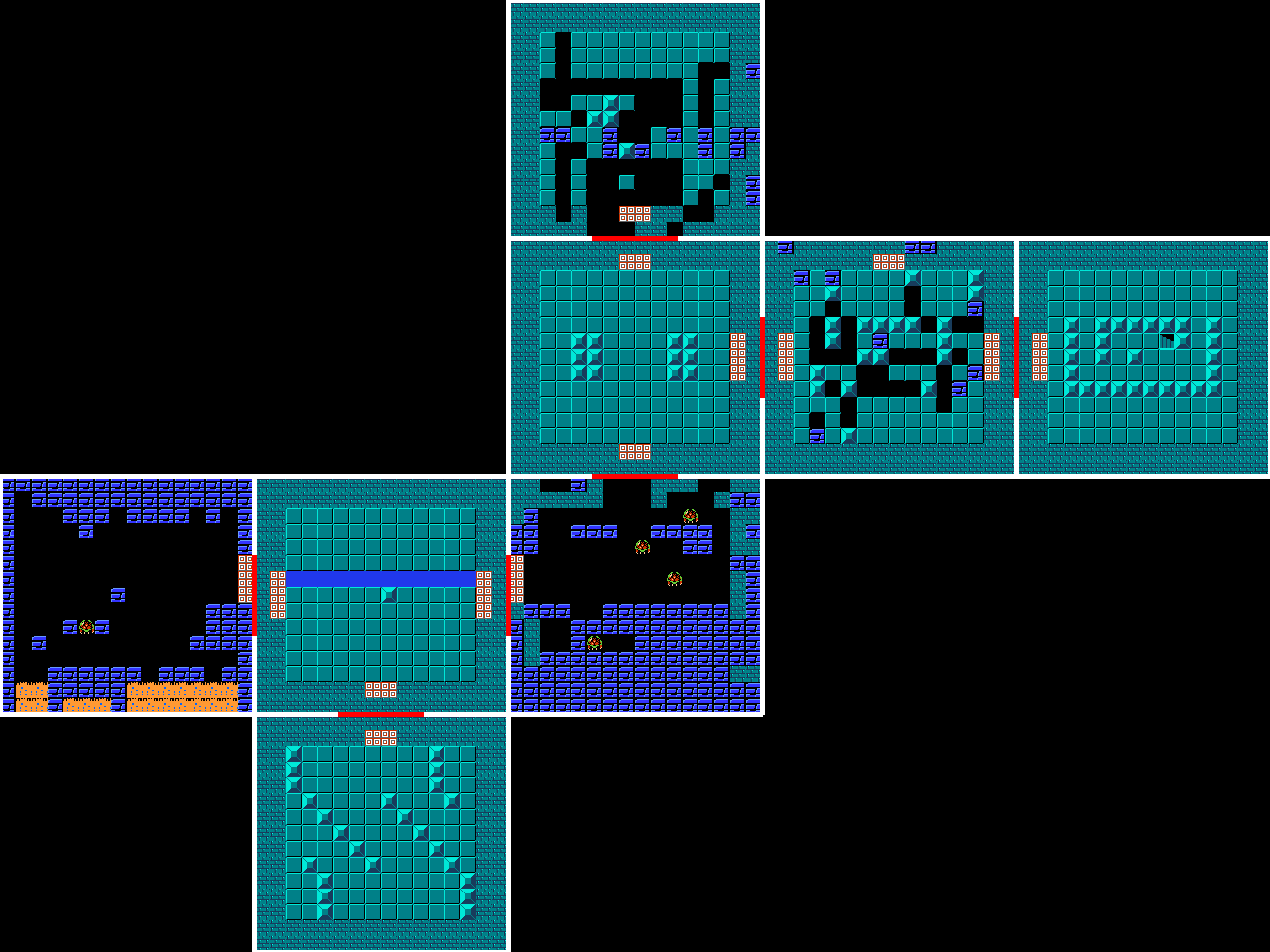}
\caption{\label{XFIGUREzeldamet} Example blended Zelda--Metroid level.}
\end{figure}
}

\newcommand{\XFIGUREzeldamm}{
\begin{figure}[t]
\centering
\includegraphics[width=1\columnwidth]{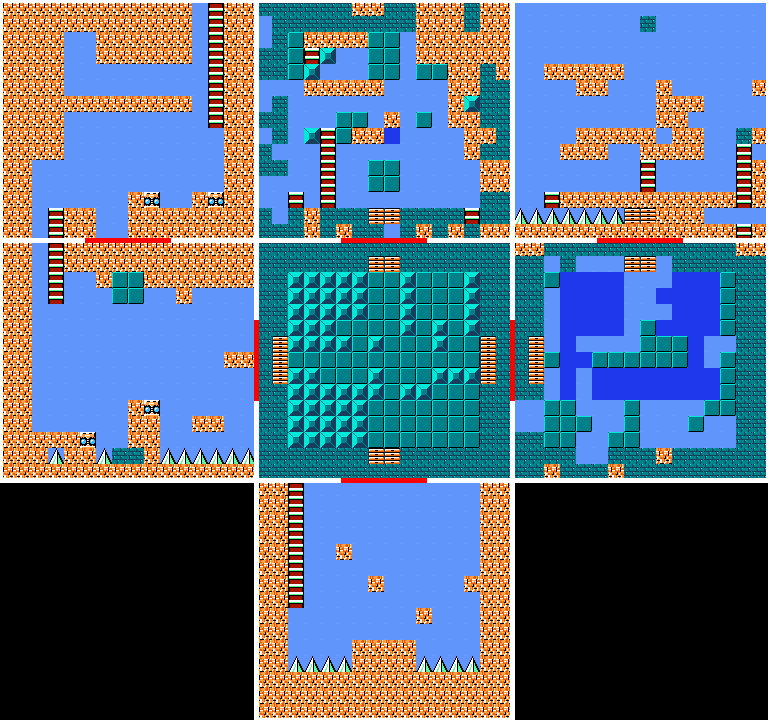}
\caption{\label{XFIGUREzeldamm} Example blended Zelda--Mega Man level.}
\end{figure}
}

\newcommand{\XFIGUREmetmm}{
\begin{figure}[t]
\centering
\includegraphics[width=1\columnwidth]{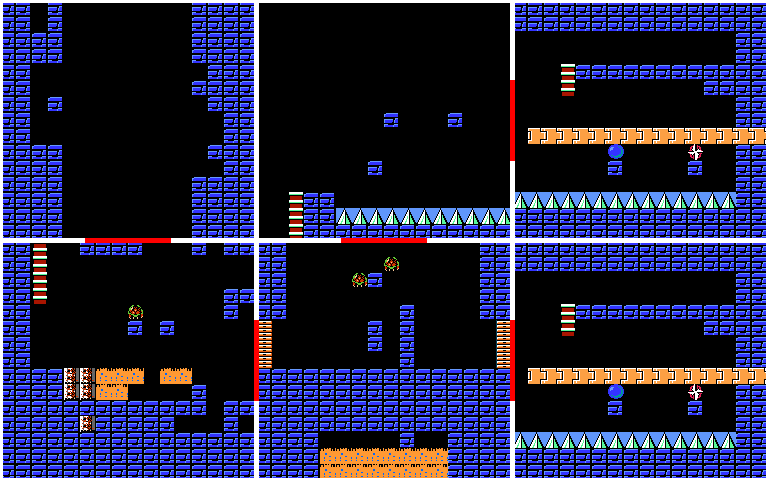}
\caption{\label{XFIGUREmetmm} Example blended Metroid--Mega Man level.}
\end{figure}
}

\newcommand{\XFIGUREzeldametmm}{
\begin{figure}[t]
\centering
\includegraphics[width=1\columnwidth]{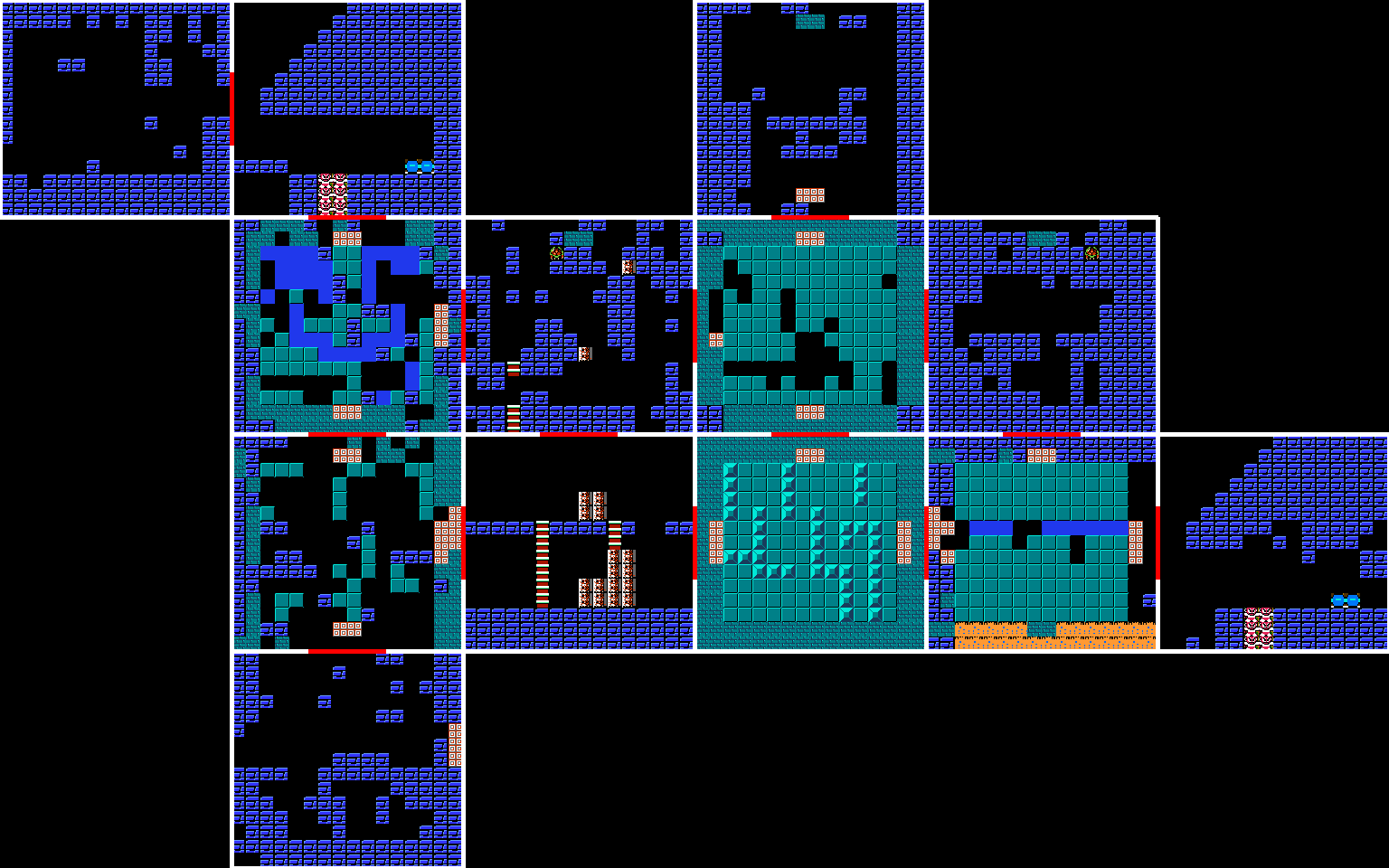}
\caption{\label{XFIGUREzeldametmm} Example blended Zelda--Metroid--Mega Man level.}
\end{figure}
}

\newcommand{\XFIGURElabels}{
\begin{figure}[t]
\centering
\begin{tabular}{cc}
\includegraphics[scale=0.25]{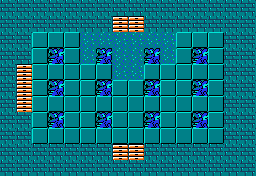} &
\includegraphics[scale=0.25]{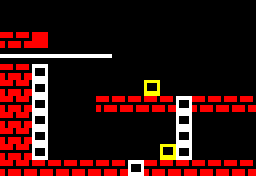}\\
Zelda - \clabel{1,1,1,0} &
Lode Runner - \clabel{0,1,1,0} \\
\\
\includegraphics[scale=0.25]{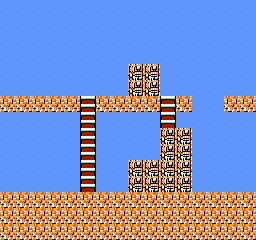} &
\includegraphics[scale=0.25]{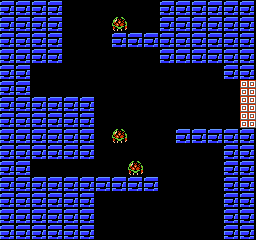} \\
Mega Man - \clabel{0,0,1,1} &
Metroid - \clabel{1,1,0,1} \\
\end{tabular}
\caption{\label{XFIGURElabels} Example segments with corresponding directional labels indicating doors/openings in \clabel{Up,Down,Left,Right}.}
\end{figure}
}

\newcommand{\XFIGUREedistance}{
\begin{figure*}[tbh!]
\centering
\begin{subfigure}[t]{0.185\textwidth}
\centering
\includegraphics[width=1\linewidth]{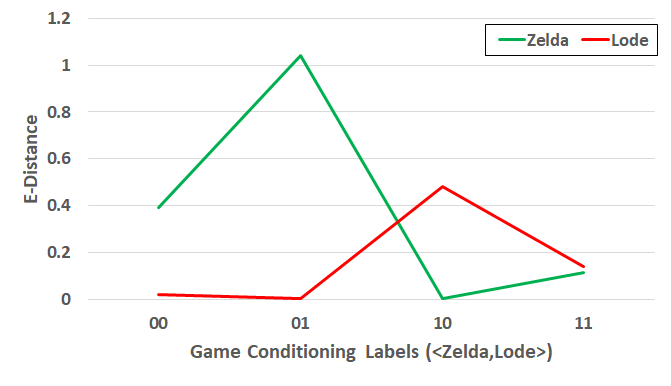}
\caption{Zelda--Lode Runner}
\end{subfigure}
~
\begin{subfigure}[t]{0.185\textwidth}
\centering
\includegraphics[width=1\linewidth]{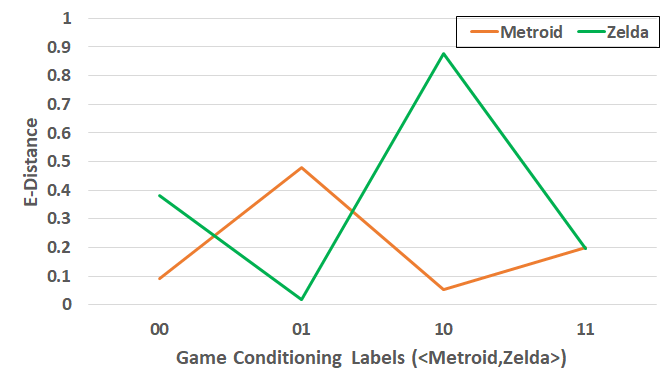}
\caption{Zelda--Metroid}
\end{subfigure}
~
\begin{subfigure}[t]{0.185\textwidth}
\includegraphics[width=1\linewidth]{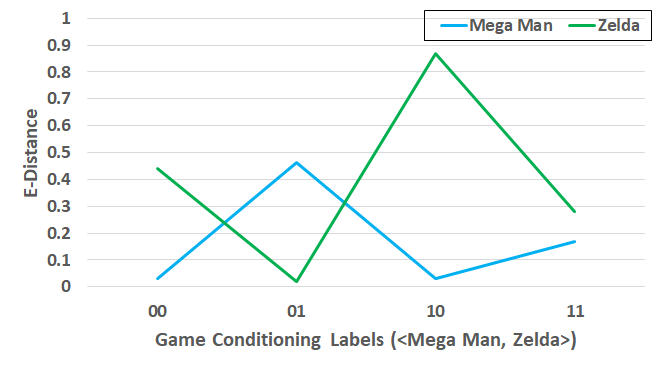}
\caption{Zelda--Mega Man}
\end{subfigure}
~
\begin{subfigure}[t]{0.185\textwidth}
\includegraphics[width=1\linewidth]{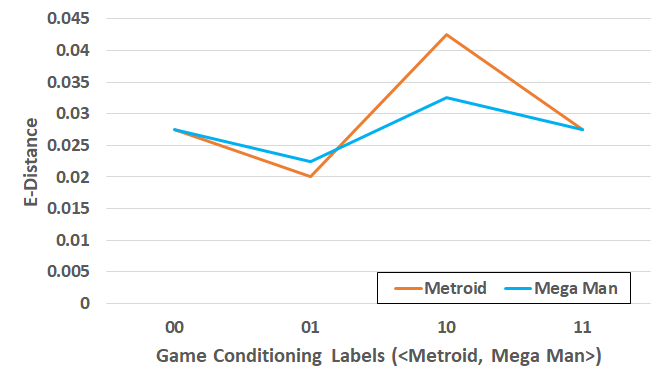}
\caption{Metroid--Mega Man}
\end{subfigure}
~
\begin{subfigure}[t]{0.185\textwidth}
\includegraphics[width=1\linewidth]{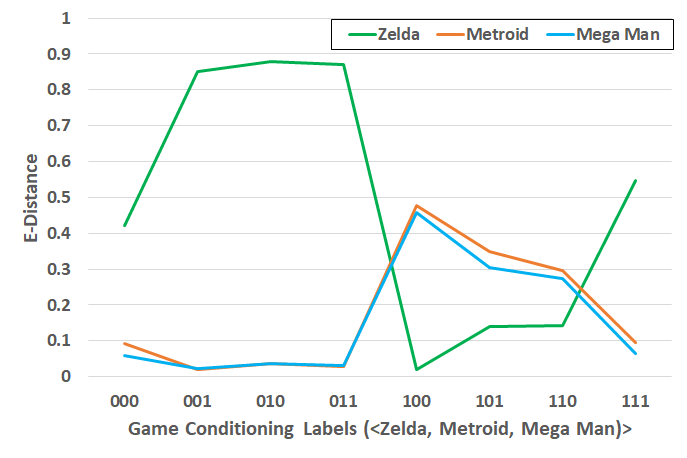}
\caption{Zelda--Met--MM}
\end{subfigure}
\caption{\label{XFIGUREedistance} E-Distances for different game blends.}
\end{figure*}
}



\newcommand{\XTABLElabelacc}{
\begin{table*}[t!]
\centering
\setlength{\tabcolsep}{2pt}
\resizebox{\textwidth}{!}{%
\begin{tabular}{|c|c|c||c|c|c||c|c|c||c|c|c|}
\hline
            & \multicolumn{2}{c||}{Zelda} & \multicolumn{3}{c||}{Metroid}  & \multicolumn{3}{c||}{Mega Man} & \multicolumn{3}{c|}{Lode Runner}     \\ \hline
Latent & Exact        & Admissible        & Exact-IN & Admissible-IN & Admissible-OUT & Exact-IN & Admissible-IN & Admissible-OUT & Exact-IN & Admissible-IN & Admissible-OUT \\ \hline
4           & 99.81        & 99.84       & \textbf{69.52}    & 90.96   & 71.15    & \textbf{61.9}     & 88.81   & 27.9     & \textbf{95.48}    & \textbf{95.48}   & 39.36    \\ \hline
8           & \textbf{99.96}        & \textbf{99.96}       & 69.28    & \textbf{91.29}   & \textbf{73.33}    & 61.62    & \textbf{89.4}    & 27.87    & 94.78    & 94.78   & 39.97    \\ \hline
16          & 99.85        & 99.91       & 67.63    & 90.38   & 72.13    & 58.9     & 86.57   & 28.14    & 94.78    & 94.78   & 39.9     \\ \hline
32          & 99.95        & \textbf{99.96}       & 62.34    & 89.33   & 64.6     & 53.78    & 83.21   & \textbf{28.99}    & 94.78    & 94.78   & \textbf{40.18}    \\ \hline
\end{tabular}
}%
\caption{\label{XTABLElabelacc} Directional label accuracy.}
\end{table*}
}

\newcommand{\XTABLEalltwoblends}{
\begin{table}[t!]
\centering
\setlength{\tabcolsep}{3pt}
\resizebox{\columnwidth}{!}{%
\begin{tabular}{|c|c|c||c|c||c|c||c|c|}
\hline
\multirow{2}{*}{\clabel{Zelda,LR}} & \multicolumn{2}{c||}{4} & \multicolumn{2}{c||}{8} & \multicolumn{2}{c||}{16} & \multicolumn{2}{c|}{32} \\ \cline{2-9} 
                       & Z          & LR        & Z          & LR        & Z          & LR         & Z          & LR         \\ \hline
\clabel{0,0}                     & 48.5       & \textbf{51.5}      & \textbf{51.6}       & 48.4      & \textbf{52.2}       & 47.8       & \textbf{62.1}       & 37.9       \\ \hline
\clabel{0,1}                     & 0          & \textbf{100}       & 0.2        & \textbf{99.8}      & 0          & \textbf{100}        & 1          & \textbf{99}         \\ \hline
\clabel{1,0}                     & \textbf{100}        & 0         & \textbf{100}        & 0         & \textbf{100}        & 0          & \textbf{100}        & 0          \\ \hline
\clabel{1,1}                     & \textbf{74.9}       & 25.1      & \textbf{95.1}       & 4.9       & \textbf{88.7}       & 11.3       & \textbf{67.3}       & 32.7       \\ \hline
\hline
\multirow{2}{*}{\clabel{Metroid,MM}} & \multicolumn{2}{c||}{4} & \multicolumn{2}{c||}{8} & \multicolumn{2}{c||}{16} & \multicolumn{2}{c|}{32} \\ \cline{2-9} 
                       & Met        & MM        & Met        & MM        & Met        & MM         & Met        & MM         \\ \hline
\clabel{0,0}                     & \textbf{53.9}       & 46.1      & \textbf{50.7}       & 49.3      & \textbf{54.1}       & 45.9       & 48.2       & \textbf{51.8}       \\ \hline
\clabel{0,1}                     & 1.4        & \textbf{98.6}      & 3.1        & \textbf{96.9}      & 5.1        & \textbf{94.9}       & 4.1        & \textbf{95.9}       \\ \hline
\clabel{1,0}                     & \textbf{96.7}       & 3.3       & \textbf{94.3}       & 5.7       & \textbf{92}         & 8          & \textbf{92.7}       & 7.3        \\ \hline
\clabel{1,1}                     & 37.3       & \textbf{62.7}      & 46.2       & \textbf{53.8}      & 44         & \textbf{56}         & 49.8       & \textbf{50.2}       \\ \hline
\hline
\multirow{2}{*}{\clabel{Metroid,Zelda}} & \multicolumn{2}{c||}{4} & \multicolumn{2}{c||}{8} & \multicolumn{2}{c||}{16} & \multicolumn{2}{c|}{32} \\ \cline{2-9} 
                       & Met        & Z         & Met        & Z         & Met        & Z          & Met        & Z          \\ \hline
\clabel{0,0}                     & \textbf{66.9}       & 33.1      & 35.9       & \textbf{64.1}      & \textbf{53.9}       & 46.1       & 47.6       & \textbf{52.4}       \\ \hline
\clabel{0,1}                     & 0          & \textbf{100}       & 0          & \textbf{100}       & 0          & \textbf{100}        & 0          & \textbf{100}        \\ \hline
\clabel{1,0}                     & \textbf{99.4}       & 0.6       & \textbf{99.7}       & 0.3       & \textbf{100}        & 0          & \textbf{99.8}       & 0.2        \\ \hline
\clabel{1,1}                     & 38.1       & \textbf{61.9}      & 31.8       & \textbf{68.2}      & 16         & \textbf{84}         & 25.9       & \textbf{74.1}       \\ \hline
\hline
\multirow{2}{*}{\clabel{MM,Zelda}} & \multicolumn{2}{c||}{4} & \multicolumn{2}{c||}{8} & \multicolumn{2}{c||}{16} & \multicolumn{2}{c|}{32} \\ \cline{2-9} 
                       & MM         & Z         & MM         & Z         & MM         & Z          & MM         & Z          \\ \hline
\clabel{0,0}                     & \textbf{64.8}       & 35.2      & 38.3       & \textbf{61.7}      & \textbf{54.9}       & 45.1       & \textbf{61.3}       & 38.7       \\ \hline
\clabel{0,1}                     & 0          & \textbf{100}       & 0          & \textbf{100}       & 0          & \textbf{100}        & 0          & \textbf{100}        \\ \hline
\clabel{1,0}                     & \textbf{98.8}       & 1.2       & \textbf{99.9}       & 0.1       & \textbf{100}        & 0          & \textbf{99.3}       & 0.7        \\ \hline
\clabel{1,1}                     & 10.8       & \textbf{89.2}      & 24.7       & \textbf{75.3}      & 41.6       & \textbf{58.4}       & \textbf{71.1}       & 28.9       \\ \hline
\end{tabular}
}%
\caption{\label{XTABLEalltwoblends} Blend accuracy for all two-game blends.}
\end{table}
}

\newcommand{\XTABLEmetmm}{
\begin{table}[t!]
\centering
\scriptsize
\setlength{\tabcolsep}{3pt}
\begin{tabular}{|c|c|c|c|c|c|c|c|c|}
\hline
\multirow{2}{*}{Label} & \multicolumn{2}{c|}{4} & \multicolumn{2}{c|}{8} & \multicolumn{2}{c|}{16} & \multicolumn{2}{c|}{32} \\ \cline{2-9} 
                       & Met        & MM        & Met        & MM        & Met        & MM         & Met        & MM         \\ \hline
00                     & \textbf{53.9}       & 46.1      & \textbf{50.7}       & 49.3      & \textbf{54.1}       & 45.9       & 48.2       & \textbf{51.8}       \\ \hline
01                     & 1.4        & \textbf{98.6}      & 3.1        & \textbf{96.9}      & 5.1        & \textbf{94.9}       & 4.1        & \textbf{95.9}       \\ \hline
10                     & \textbf{96.7}       & 3.3       & \textbf{94.3}       & 5.7       & \textbf{92}         & 8          & \textbf{92.7}       & 7.3        \\ \hline
11                     & 37.3       & \textbf{62.7}      & 46.2       & \textbf{53.8}      & 44         & \textbf{56}         & 49.8       & \textbf{50.2}       \\ \hline
\end{tabular}
\caption{\label{XTABLEmetmm} Metroid--Mega Man}
\end{table}
}

\newcommand{\XTABLEzelmet}{
\begin{table}[t!]
\centering
\scriptsize
\setlength{\tabcolsep}{3pt}
\begin{tabular}{|c|c|c|c|c|c|c|c|c|}
\hline
\multirow{2}{*}{Label} & \multicolumn{2}{c|}{4} & \multicolumn{2}{c|}{8} & \multicolumn{2}{c|}{16} & \multicolumn{2}{c|}{32} \\ \cline{2-9} 
                       & Met        & Z         & Met        & Z         & Met        & Z          & Met        & Z          \\ \hline
00                     & \textbf{66.9}       & 33.1      & 35.9       & \textbf{64.1}      & \textbf{53.9}       & 46.1       & 47.6       & \textbf{52.4}       \\ \hline
01                     & 0          & \textbf{100}       & 0          & \textbf{100}       & 0          & \textbf{100}        & 0          & \textbf{100}        \\ \hline
10                     & \textbf{99.4}       & 0.6       & \textbf{99.7}       & 0.3       & \textbf{100}        & 0          & \textbf{99.8}       & 0.2        \\ \hline
11                     & 38.1       & \textbf{61.9}      & 31.8       & \textbf{68.2}      & 16         & \textbf{84}         & 25.9       & \textbf{74.1}       \\ \hline
\end{tabular}
\caption{\label{XTABLEzelmet} Zelda--Metroid}
\end{table}
}

\newcommand{\XTABLEzelmm}{
\begin{table}[t!]
\centering
\scriptsize
\setlength{\tabcolsep}{3pt}
\begin{tabular}{|c|c|c|c|c|c|c|c|c|}
\hline
\multirow{2}{*}{Label} & \multicolumn{2}{c|}{4} & \multicolumn{2}{c|}{8} & \multicolumn{2}{c|}{16} & \multicolumn{2}{c|}{32} \\ \cline{2-9} 
                       & MM         & Z         & MM         & Z         & MM         & Z          & MM         & Z          \\ \hline
00                     & \textbf{64.8}       & 35.2      & 38.3       & \textbf{61.7}      & \textbf{54.9}       & 45.1       & \textbf{61.3}       & 38.7       \\ \hline
01                     & 0          & \textbf{100}       & 0          & \textbf{100}       & 0          & \textbf{100}        & 0          & \textbf{100}        \\ \hline
10                     & \textbf{98.8}       & 1.2       & \textbf{99.9}       & 0.1       & \textbf{100}        & 0          & \textbf{99.3}       & 0.7        \\ \hline
11                     & 10.8       & \textbf{89.2}      & 24.7       & \textbf{75.3}      & 41.6       & \textbf{58.4}       & \textbf{71.1}       & 28.9       \\ \hline
\end{tabular}
\caption{\label{XTABLEzelmm} Zelda--Mega Man}
\end{table}
}

\newcommand{\XTABLEzelmetmm}{
\begin{table}[t!]
\centering
\setlength{\tabcolsep}{3pt}
\resizebox{\columnwidth}{!}{%
\begin{tabular}{|c|c|c|c||c|c|c||c|c|c||c|c|c|}
\hline
\multirow{2}{*}{\shortstack{\clabelopen Zelda,\\[-2pt]Met,MM\clabelclose}} & \multicolumn{3}{c||}{4}                        & \multicolumn{3}{c||}{8}               & \multicolumn{3}{c||}{16}                       & \multicolumn{3}{c|}{32}                       \\ \cline{2-13} 
                       & Zel           & Met           & MM            & Zel           & Met  & MM            & Zel           & Met           & MM            & Zel           & Met           & MM            \\ \hline
\clabel{0,0,0}                    & \textbf{54.1} & 23.3          & 22.6          & \textbf{38.2} & 33.1 & 28.7          & 34.8          & \textbf{36.3} & 28.9          & \textbf{54.2} & 25.3          & 20.5          \\ \hline
\clabel{0,0,1}                    & 0.3           & 8.5           & \textbf{91.2} & 2.4           & 5.8  & 91.8          & 1             & 9.9           & \textbf{89.1} & 0.1           & 13.3          & \textbf{86.6} \\ \hline
\clabel{0,1,0}                    & 0.1           & \textbf{94.1} & 5.8           & 0             & 91.3 & 8.7           & 0.2           & \textbf{94}   & 5.8           & 0             & \textbf{89.1} & 10.9          \\ \hline
\clabel{0,1,1}                    & 0.1           & 41.1          & \textbf{58.8} & 0.1           & 38.8 & \textbf{61.1} & 0             & 47.2          & \textbf{52.8} & 0             & 46.9          & \textbf{53.1} \\ \hline
\clabel{1,0,0}                    & \textbf{100}  & 0             & 0             & \textbf{100}  & 0    & 0             & \textbf{100}  & 0             & 0             & \textbf{100}  & 0             & 0             \\ \hline
\clabel{1,0,1}                    & \textbf{91.2} & 0.5           & 8.3           & \textbf{89.8} & 0.3  & 9.9           & \textbf{98.2} & 0.3           & 1.5           & \textbf{51.3} & 6.1           & 42.6          \\ \hline
\clabel{1,1,0}                    & \textbf{75}   & 21.6          & 3.4           & \textbf{90.9} & 8.7  & 0.4           & \textbf{96.8} & 3             & 0.2           & \textbf{61.8} & 31.3          & 6.9           \\ \hline
\clabel{1,1,1}                    & 23.6          & 32.9          & \textbf{43.5} & 44.4          & 21.1 & \textbf{34.5} & \textbf{48.2} & 24.7          & 27.1          & 6.5           & 33.3          & \textbf{60.2} \\ \hline
\end{tabular}
}%
\caption{\label{XTABLEzelmetmm} Blend Accuracy for Zelda--Metroid--Mega Man.}
\end{table}
}

\newcommand{\XTABLEalldensym}{
\begin{table}[t!]
\centering
\setlength{\tabcolsep}{2pt}
\resizebox{\columnwidth}{!}{%
\begin{tabular}{|c|c||c|c|c|c|}
\hline
\textbf{Zelda} & Original    & 4           & 8           & 16          & 32          \\ \hline
Density  & $0.59\pm0.08$ & $0.58\pm0.06$ & $0.58\pm0.06$ & $\mathbf{0.59\pm0.06}$ & $0.59\pm0.07$ \\ \hline
Symmetry & $0.76\pm0.06$ & $0.76\pm0.04$ & $0.76\pm0.04$ & $0.76\pm0.05$ & $0.75\pm0.05$ \\ \hline
\hline
\textbf{Metroid} & Original    & 4           & 8          & 16          & 32          \\ \hline
Density  & $0.39\pm0.16$ & $\mathbf{0.41\pm0.09}$ & $\mathbf{0.4\pm0.08}$ & $\mathbf{0.41\pm0.08}$ & $\mathbf{0.4\pm0.08}$  \\ \hline
Symmetry & $0.67\pm0.12$ & $\mathbf{0.59\pm0.04}$ & $\mathbf{0.6\pm0.05}$ & $\mathbf{0.6\pm0.05}$  & $\mathbf{0.61\pm0.05}$ \\ \hline
\hline
 \textbf{Mega Man} & Original    & 4           & 8           & 16          & 32          \\ \hline
Density  & $0.38\pm0.18$ & $0.4\pm0.13$  & $0.41\pm0.13$ & $0.39\pm0.14$ & $0.39\pm0.13$ \\ \hline
Symmetry & $0.65\pm0.13$ & $0.62\pm0.08$ & $0.63\pm0.07$ & $0.64\pm0.09$ & $0.63\pm0.07$ \\ \hline
\hline
 \textbf{Lode} & Original    & 4           & 8           & 16          & 32          \\ \hline
Density     & $0.41\pm0.17$ & $0.39\pm0.14$ & $0.39\pm0.15$ & $\mathbf{0.38\pm0.15}$ & $0.39\pm0.15$ \\ \hline
Symmetry    & $0.52\pm0.14$ & $0.51\pm0.1$  & $0.53\pm0.1$  & $\mathbf{0.53\pm0.11}$ & $\mathbf{0.53\pm0.11}$ \\ \hline
\end{tabular}
}%
\caption{\label{XTABLEalldensym} Means $\pm$ std. devs. for \textit{Density} and \textit{Symmetry} of generated segments. Values in bold were significantly different from the original segments.}
\end{table}
}

\newcommand{\XTABLEmetdensym}{
\begin{table}[t!]
\scriptsize
\centering
\setlength{\tabcolsep}{2pt}
\begin{tabular}{|c|c|c|c|c|c|}
\hline
  & Original    & 4           & 8          & 16          & 32          \\ \hline
Density  & $0.39\pm0.16$ & $0.41\pm0.09$ & $0.4\pm0.08$ & $0.41\pm0.08$ & $0.4\pm0.08$  \\ \hline
Symmetry & $0.67\pm0.12$ & $0.59\pm0.04$ & $0.6\pm0.05$ & $0.6\pm0.05$  & $0.61\pm0.05$ \\ \hline
\end{tabular}
\caption{\label{XTABLEmetdensym} Metroid Den Sym}
\end{table}
}

\newcommand{\XTABLEmmdensym}{
\begin{table}[t!]
\scriptsize
\centering
\setlength{\tabcolsep}{2pt}
\begin{tabular}{|c|c|c|c|c|c|}
\hline
 & Original    & 4           & 8           & 16          & 32          \\ \hline
Density  & $0.38\pm0.18$ & $0.4\pm0.13$  & $0.41\pm0.13$ & $0.39\pm0.14$ & $0.39\pm0.13$ \\ \hline
Symmetry & $0.65\pm0.13$ & $0.62\pm0.08$ & $0.63\pm0.07$ & $0.64\pm0.09$ & $0.63\pm0.07$ \\ \hline
\end{tabular}
\caption{\label{XTABLEmmdensym} MM Den Sym}
\end{table}
}

\newcommand{\XTABLElodedensym}{
\begin{table}[t!]
\scriptsize
\centering
\setlength{\tabcolsep}{2pt}
\begin{tabular}{|c|c|c|c|c|c|}
\hline
 & Original    & 4           & 8           & 16          & 32          \\ \hline
Density     & $0.41\pm0.17$ & $0.39\pm0.14$ & $0.39\pm0.15$ & $0.38\pm0.15$ & $0.39\pm0.15$ \\ \hline
Symmetry    & $0.52\pm0.14$ & $0.51\pm0.1$  & $0.53\pm0.1$  & $0.53\pm0.11$ & $0.53\pm0.11$ \\ \hline
\end{tabular}
\caption{\label{XTABLElodedensym} Lode Den Sym}
\end{table}
}

\newcommand{\XTABLEnovelty}{
\begin{table}[t!]
\centering
\setlength{\tabcolsep}{3pt}
\resizebox{\columnwidth}{!}{%
\begin{tabular}{|c|c||c|c|c||c|c|c||c|c|c|}
\hline
\multirow{2}{*}{Latent} & Zelda         & \multicolumn{3}{c||}{Metroid}                  & \multicolumn{3}{c||}{Mega Man}                 & \multicolumn{3}{c|}{Lode Runner}              \\ \cline{2-11} 
                  & Ovr           & IN            & OUT           & Ovr           & IN            & OUT           & Ovr           & IN            & OUT           & Ovr           \\ \hline
4                 & 43.1          & \textbf{59.9} & \textbf{88.8} & \textbf{67.2} & 69.8          & \textbf{95.9} & 81.5          & 78.4          & \textbf{99.5} & 94.2          \\ \hline
8                 & 45.2          & 58.1          & 87.3          & 66.6          & \textbf{72.7} & 95.4          & \textbf{83.3} & 82.6          & 99.2          & 94.7          \\ \hline
16                & 45.8          & 58.4          & 86.6          & 65.5          & 69.9          & 93.6          & 80            & \textbf{84.2} & 98.8          & \textbf{95.8} \\ \hline
32                & \textbf{48.3} & 58.6          & 77.1          & 63.6          & 71.9          & 89.1          & 79.3          & 84.1          & 98.9          & 95.5          \\ \hline
\end{tabular}
}
\caption{\label{XTABLEnovelty} Novelty of generated segments. \emph{Ovr} is the overall percentage, combining both IN and OUT label percentages.}
\end{table}
}



\sectionpre
\section{Introduction}
\sectionpost
Among the many approaches for Procedural Content Generation via Machine Learning (PCGML) \cite{summerville2017procedural}, variational autoencoders (VAEs) \cite{kingma2013autoencoding} have been used in several recent works for generating levels \cite{yang2020game,thakkar2019autoencoder} as well as blending levels across different games \cite{sarkar2019blending,sarkar2020exploring}. VAEs consist of encoder and decoder networks that learn to map data to and from a continuous latent space respectively. Once trained, the latent space can be sampled to generate new data. While shown to be effective in generating and blending levels for several games, standard VAEs work with fixed size inputs/outputs, thus forcing models to train on and thereby generate level segments rather than whole levels, due to the dearth of training data for games. Additionally, VAEs enable generation via random sampling of latent space vectors, thus methods such as latent variable evolution \cite{bontrager2018deepmasterprints} have been used to add controllability by searching for latent vectors that satisfy certain properties. Alternatively, conditional VAEs \cite{sohn2015learning} augment regular VAEs by allowing them to be trained using labeled data and thus enable the generation of outputs conditioned by specific labels. This obviates having to run evolutionary search post-training and enables controllability via the training process itself. Prior work \cite{sarkar2020conditional} explored the use of CVAEs for PCGML, finding them capable of generating outputs with desired elements and patterns as well as blending outputs across games. However, this was limited to level segments and like much PCGML research, was restricted to the platformer domain.

In this paper, we expand on the above work by using conditional VAEs to generate entire levels and do so for multiple genres. Specifically, we show that by having the conditioning labels indicate the direction of door placement in dungeon rooms and progression in platformer level segments, we can generate rooms and segments with orientations that enable them to be reliably connected in order to form whole levels. Moreover, this also enables us to generate levels consisting of both dungeon rooms and platformer level segments, thus enabling blending levels from different genres unlike prior PCGML blending work which did so only across different platformers. We demonstrate our approach using levels from the platformers \textit{Metroid} and \textit{Mega Man}, the puzzle platformer \textit{Lode Runner} and dungeons from \textit{The Legend of Zelda}. Our work thus contributes, to our knowledge, 1) a new PCGML approach for generating whole levels using conditional VAEs and the first to be applied in multiple genres and 2) the first PCGML approach for blending levels across different genres.


\XFIGURElabels

\sectionpre
\section{Background}
\sectionpost
A significant body of PCGML research has used variational autoencoders (VAEs) \cite{kingma2013autoencoding} with most focusing on generating and blending platformer levels \cite{sarkar2019blending,sarkar2020exploring,yang2020game,sarkar2020conditional}. Advanced models such as Gaussian Mixture VAEs \cite{yang2020game} have also been used for injecting controllability into vanilla VAE models. Controllability has been the focus of several PCGML works involving GANs \cite{goodfellow2014generative} and VAEs since random sampling of the latent space does not afford much control. A common approach for controllability is latent variable evolution \cite{bontrager2018deepmasterprints} which utilizes evolution to look for latent vectors corresponding to desired content, as demonstrated for both GANs \cite{volz2018evolving} and VAEs \cite{sarkar2019blending}. Relatedly, \cite{schrum2020interactive} enabled interactive exploration of a GAN latent space to evolve desired Mario levels and Zelda dungeons. 

Unlike these hybrid approaches that add controllability via a method separate from training the model, conditional VAEs (CVAEs) \cite{sohn2015learning} enable control as part of the model itself. When training CVAEs, each input is associated with a user-specified label. The encoder and decoder then learn to use these labels, and thus the information provided by them, to map inputs to and from the latent space respectively. Prior work \cite{sarkar2020conditional} showed that CVAEs can generate levels conditioned on labels specifying the content and patterns desired to be present in the levels as well as blend levels from different games by having the label indicate the games to blend. We build on this by expanding beyond platformers and generating whole levels. 

Generating whole levels is particularly challenging when working with latent models for two reasons. First, such models work with fixed size inputs and outputs, and second, the general dearth of training data for PCGML necessitating training on segments rather than whole levels to ensure enough training data. While generating whole levels by stitching together successively generated segments \cite{green2020mario} is acceptable for a game like Mario with uniform progression, this does not work for games that progress in multiple directions and orientations. This was addressed in \cite{sarkar2020sequential} via an approach combining a VAE-based sequential model and a classifier to generate and then place a segment relative to the previous one, thus generating whole levels by an iterative loop of decoding and encoding successive segments. However this afforded control only via definition of the initial segment with the orientation of successive segments and properties of the blend not being controllable. This issue was also addressed by \cite{capps2021using} which focused on generating whole Mega Man levels by using multiple GANs to model levels with different orientations. Our work differs in that compared to \cite{sarkar2020sequential}, each segment and the blend properties are controllable and compared to \cite{capps2021using}, we use VAEs and model differences in orientations with separate labels rather than separate models. Additionally, we extend beyond the platformer domain.

While most PCGML works have focused on side-scrolling platformers, a handful have worked with other genres. Doom levels have been generated using GANs\cite{giacomello2018doom} while Lode Runner levels have been generated using GANs \cite{steckel2021illuminating}, Markov models \cite{snodgrass2017learning} and VAEs \cite{thakkar2019autoencoder}. Both  \cite{summerville2015samplinghyrule} and  \cite{summerville2015learning} generated Zelda dungeons using Bayes nets. Similar to our use of latent models to create whole dungeons out of generated rooms, \cite{gutierrez2020generative} used an approach that utilized a graph grammar to generate a dungeon layout and a GAN to generate rooms to fill that layout. Our work differs in using VAEs and that the properties of  dungeon rooms are determined by conditioning rather than a grammar.

Game blending was proposed by \cite{gow2015towards} and refers to generating new games by blending levels and/or mechanics of existing games. Since then, many works \cite{sarkar2019blending,sarkar2020exploring,sarkar2020conditional,sarkar2021generating} have used VAEs for blending levels across several platformers though only working with segments. The approach in \cite{sarkar2020sequential} generated whole levels that are blended but without affording control over blend contents. The method presented in this paper allows generating whole level blends in a controllable manner in addition to blending games from different genres. In doing so, this work constitutes another PCGML approach under combinational creativity \cite{boden2004creative}, similar to recent works looking at domain transfer \cite{snodgrass2020multi} and automated game generation \cite{guzdial2021conceptual}, all of which blend and/or combine existing domains to generate content for new ones. Additionally, in being a creative ML approach for level generation and blending, this work can also be categorized under the recently defined Game Design via Creative Machine Learning (GDCML) \cite{sarkar2020towards} set of approaches.

\sectionpre
\section{Method}
\sectionpost

\subsection{Level Representation and Conditioning}
\subsectionpost
We used levels from the action-adventure game \textit{The Legend of Zelda} (Zelda), the action platformers \textit{Metroid} and \textit{Mega Man} (MM), and the puzzle platformer \textit{Lode Runner} (LR), thus spanning several genres and types of games. Levels were taken from the Video Game Level Corpus (VGLC) \cite{summerville2016vglc} and were in text format with different characters mapping to different in-game tiles. The premise that facilitates the use of CVAEs is that levels in these games are composed of segments connected together in a meaningful way to enable gameplay. This is most evident in Zelda where discrete rooms are connected based on the location of doors but is also true for platformers where levels can be considered to consist of discrete segments connected based on the directionality of player movement through them. Thus, training a CVAE on the discrete segments while using labels indicating how they can be connected to other segments should yield a model that can generate whole levels composed of these interconnected segments. For each game, labels were binary-encoded vectors of length 4 with each element corresponding to a direction with 1/0 indicating that progression in that direction was possible/not possible. For all games, the order was Up, Down, Left, Right, thus, a label of \clabel{1,0,0,1} indicated that progression was possible in the upward and rightward, but not in the downward or leftward, direction for the corresponding segment.

\subsubsection{The Legend of Zelda}
Training data for Zelda consisted of the discrete rooms taken from each dungeon in the VGLC. Rooms in Zelda are 11x16 tiles consisting of a 2-tile perimeter of wall tiles and a 7x12 playing area. We augmented the training set by adding horizontally and vertically flipped versions of each room, if not already present, finally ending up with a total of 502 rooms for training. Labels for each room were determined based on the location of the door tiles.

\subsubsection{Metroid}
For Metroid, we extracted 15x16 segments from the whole levels. Though Metroid levels are not composed of independent rooms like Zelda dungeons, we could still extract discrete segments based on the fact that horizontal segments of Metroid are 15 tiles high and vertical segments are 16 tiles wide. After filtering out segments that consisted entirely of block characters, we obtained 414 segments. Labels for each segment were assigned manually based on visual inspection and indicated in which directions that segment was open i.e. in which directions could a player enter or exit that segment. Doors in Metroid were considered an open direction.

\subsubsection{Mega Man}
Horizontal and vertical sections of MM have similar dimensions as in Metroid so here too, we extracted 15x16 segments, obtaining 143 for training. Labels were assigned similarly to how they were for Metroid segments.

\subsubsection{Lode Runner}
All LR levels have a fixed dimension of 22x32 tiles. Thus, a standard VAE is perfectly suited to generate whole levels but we still wanted to test our CVAE approach of building levels using labeled segments with LR and thus divided each of the 150 LR levels in the VGLC into 4 segments of 11x16 tiles, obtaining a total of 600 segments for training. Labels for each segment were assigned to indicate in which direction that segment was connected to another e.g. the top-left segment was assigned \clabel{0,1,0,1} i.e. Down and Right, since it had a segment to the right and a segment below it.

Unlike prior work using VAEs, rather than extract segments with redundancy from platformer levels by sliding a window one column/row at a time, we used non-overlapping segments to be more compatible with the discrete Zelda rooms as well as for directional labels to be more meaningful and avoid situations such as e.g., Metroid segments with doors in the middle. Example segments and labels are shown in Figure \ref{XFIGURElabels}. 

\subsectionpre
\subsection{Game and Genre Blending}
\subsectionpost
In addition to working with the games separately, we wanted to test the CVAE in generating levels blended across different games. Since the labels indicate directions in which progress is possible, irrespective of whether they are doors in dungeon rooms or pathways/openings in platformers, a model trained on similarly labeled sections from multiple games taken together might be able to produce blended levels composed of such interconnected sections. For example, a model trained on Zelda and Metroid should be able to produce blended levels composed of Zelda rooms connected with Metroid segments and blended segments consisting of Zelda and Metroid elements, thus giving us an approach for potentially blending levels across not just games but different genres. For these CVAE models, we extended the labels to also specify the games being blended i.e. for a model trained on segments/rooms from $n$ games, we concatenated an $n$-element binary vector to the 4-element binary vector indicating the directionality as before e.g. when blending two games, labels for segments from game 1 were concatenated with \clabel{1,0} while those from game 2 were concatenated with \clabel{0,1}. We trained blended models of \textit{Metroid--Mega Man}, \textit{Zelda--Lode Runner}, \textit{Zelda--Metroid}, \textit{Zelda--Mega Man} and \textit{Zelda--Mega Man--Metroid}.

Training a blended model on Metroid and MM was straightforward since both consist of 15x16 segments. Similarly, 11x16 Zelda rooms were compatible with 11x16 LR segments. For blending Zelda with Metroid and MM, we padded the Zelda rooms by duplicating each room's two outermost floor rows both to the north and to the south, thus yielding 15x16 rooms. Being discrete and independent, Zelda rooms were well-suited to being padded. Padding LR segments was less resistant to impacting directionality. Thus for that and to keep to a manageable number of models, we blended LR with only Zelda. For the 3-game blend, game labels were \clabel{0,0,1}, \clabel{0,1,0} and \clabel{1,0,0} referring to MM, Metroid and Zelda respectively. 

Besides using models trained on multiple games taken together, blending can also be done by taking turns generating level sections using multiple single-game models. This was explored in \cite{sarkar2018blending} using separate LSTM models of Mario and \textit{Kid Icarus}. Since our CVAEs use the same underlying definition for labels (i.e. directionality), given a set of labels, we can use different models to generate different level sections to obtain a level made of segments from multiple games, as in Figure 1. One could select which game model to use based on specified probabilities which enables generating levels blending desired proportions of games. Since this uses
the single-game models, we do not separately evaluate this approach.

\XTABLElabelacc
\XTABLEalltwoblends
\XTABLEzelmetmm

\subsectionpre
\subsection{Layout Algorithm}
\subsectionpost
We use a simple approach to generate layouts within which to place segments. We start by placing a single segment with closed labels on all four sides. At each step, we select a random side of a segment labeled as closed. If there is no segment next to that side, a segment is placed there with all sides labeled closed. We then connect the originally selected segment to the (pre-existing or newly placed) segment next to the selected side by setting the label to open on the adjoining sides. The number of steps to run is chosen randomly from a small range. We combine this with the CVAE by looping through each layout segment, determining the appropriate label based on the directions in which that segment is open/closed and then use that to condition the generation of a sampled latent vector to place in that location. For blending, any provided game label could be concatenated to the determined direction label to produce a segment. For the blend examples shown in the paper, we randomly set the game label bits to 1 or 0 for each game. This method could also accept probabilities for each game and then generate levels that blend the games accordingly.

\subsectionpre
\subsection{Modeling and Conditional Training/Generation}
\subsectionpost
We trained a CVAE model for each of the above games and blends. The encoder and decoder for all models consisted of 4 fully connected layers each, using ReLU activation. Models were trained for 10000 epochs using the Adam optimizer and a learning rate of 0.001, decayed by 0.01 every 2500 epochs. These hyperparameters were set based on the prior VAE-based PCGML work. All models were trained using PyTorch \cite{paszke2017automatic}. To study different latent sizes, we trained 4 versions of each model using latent dimensions of 4, 8, 16 and 32. 

\XTABLEalldensym
\XTABLEnovelty

Training a CVAE involves associating a label with each input, concatenating them together and forwarding them through the encoder to obtain a latent vector which is then concatenated with the same label and forwarded through the decoder. Through training, the encoder and decoder thus learn to use the label to encode inputs and generate outputs respectively. After training, generation is done by sampling a vector from the latent space, concatenating it with the desired label and forwarding it through the decoder to obtain the output. More technical details regarding CVAEs can be found in \cite{sohn2015learning,yan2015conditional}.

\XFIGUREedistance

\sectionpre
\section{Evaluations and Discussion}
\sectionpost

\subsection{Directional Labeling Accuracy}
\subsectionpost
For evaluations, we were primarily interested in testing the accuracy of labeling, i.e. if the segments generated by the CVAEs were true to the directionality/orientation as indicated by the labels used to generate them. For Zelda, this is straightforward to verify since we just check the location of door tiles in the generated segments, similar to how the training segments were labeled. For each of the 3 other games, we trained a random forest classifier on their training segments using 10-fold cross validation and the directionality labels as the class labels with Metroid, MM and LR having 12, 9 and 4 unique classes/directionalities respectively. All classifiers were implemented using scikit-learn \cite{scikit-learn}. \review{We obtained mean classification accuracies of 98\%, 81\% and 51\% respectively. The low accuracy for LR is likely due to training on level quadrants and that structure and progression doesn't particularly vary across them since original levels do not vary gameplay based on quadrants. We note however that this implies that different 11x16 LR segments are interchangeable in terms of being able to be combined with respect to directionality and progression}. To evaluate conditioning accuracy, for a generated segment, we compared the label predicted by the classifier with the one used to condition the segment's generation. While ideally we want exact matches between predicted and conditioning labels, i.e. the generated segment to be open and closed in the exact label-specified directions, it is sufficient that the generated segment has just the desired open directions, regardless of whether the desired closed directions end up being open in the generated segment. We want to avoid the reverse case i.e. desired open directions being closed in the generated segment. Thus, in addition to \textit{exact} label matches, we were interested in matches where the bits set to 1 in the conditioning label were also set to 1 in the predicted label, regardless of the value of the other bits. We refer to these as \textit{admissible} matches. Note that by definition, an exact match is also admissible. An additional point of note is that not all possible directional labels are present in each game. Since the directional labels are binary vectors of length 4, there are $2^4=16$ possible labels. However, as mentioned above, only 12, 9 and 4 of these are present in the training data of Metroid, MM and LR respectively, with only Zelda having all 16. It is thus reasonable to expect that the models will do a better job of generating matching segments when using labels that appear in the training data compared to those that do not. Thus, except for Zelda, we track matches separately for IN-game and OUT-of-game labels.

For each model, we sampled 1000 latent vectors at random and used each of the 16 possible labels for conditioning, thus generating 16000 segments. For each segment, we applied the relevant classifier and compared the predicted label with the label used for conditioning, separately computing the percentage of \textit{exact} and \textit{admissible} matches for IN and OUT labels. That is, we computed what percentage of IN labels yielded \textit{exact} and \textit{admissible} matches and what percentage of OUT labels yielded such matches. Results are shown in Table \ref{XTABLElabelacc}. All Zelda models achieved near 100\% accuracy, suggesting that door placement using conditioning is very reliable for Zelda rooms. The \textit{exact} match percentages for OUT labels were 0\% in all non-Zelda games so we omitted them. The next highest percentages are for LR though only 4 out of 16 labels are IN labels so for the remaining 12, LR models produced \textit{admissible} matches only about 40\% of the time. Metroid outperformed MM in all 3 types of matches. Overall, in all cases, we observed about 90\% \textit{admissible} matches when using IN labels suggesting that the models can very reliably produce openings/doors in desired directions as prescribed by the label, even if \textit{exact} matches are not as frequent.

\subsectionpre
\subsection{Blend Labeling Accuracy}
\subsectionpost
Similarly, we wanted to evaluate the blend labeling accuracy. For each blended model, we trained a random forest classifier with 10-fold cross validation, using the game that a training segment belongs to as the class label. The classifier for the \textit{Zelda--Metroid--Mega Man} blend worked with 3 class labels while the others all worked with two. \review{For all blends involving Zelda, we obtained classification accuracies of 100\% and an accuracy of 98\% for the Metroid-Mega Man blend.}

For each model, we sampled 100 latents and used each possible game+directional label to condition their generation. For 2-game blends, there were $2^{(2+4)}$=$64$ such labels, resulting in a total of 6400 generated segments and for the 3-game blend, there were $2^{(3+4)}$=$128$ possible labels, resulting in 12800 generated segments. In both cases, we applied the relevant blend classifier on each generated segment and compared the predicted label with the game label used for generation. In all cases, for each possible game label, we computed the percentage of times each of the games in that blend was predicted. We expect that when using single game labels (e.g. \clabel{0,1}, \clabel{1,0}) to generate segments, predictions will be high for that particular game. Also, when using blended game labels (e.g. \clabel{1,1}, \clabel{1,0,1}), predictions should be more spread out since here segments would blend the properties of the original games. Results are given in Tables \ref{XTABLEalltwoblends} and \ref{XTABLEzelmetmm} and are true to expectation. In all cases, when using single-game blend labels, the classifier predicts the relevant game close to 100\% of the time for 2-game blends involving Zelda and at least 92.7\% of the time for the Metroid-MM blend. In the 3-game blend this drops slightly to a minimum of 86.6\% but is still mostly a very high percentage in all cases. Also, blend labels in the 2-game cases (\clabel{0,0}, \clabel{1,1}) lead to more spread out predictions with behavior varying across models, though in no clear pattern. There is more of a spread when using \clabel{0,0} than \clabel{1,1} with the latter in some cases causing one of the games to dominate. For the 3-game case, we also note more spread out predictions for the blend labels and note that when a game's label element is set to 0, that game is mostly predicted less than 1\% of the time, except in the \clabel{0,0,0} case. Overall, these results, combined with those for the directional labels, suggest that conditioning can successfully generate segments with desired orientations as well as blend desired games together.

\subsectionpre
\subsection{Tile-based Metrics and Novelty}
\subsectionpost
We also evaluated the actual content of the generated segments. For this, we used the following two tile-based metrics:
\begin{itemize}
    \item \textit{Density} - proportion of a segment/room that is walls, ground/breakable tiles, doors, platforms and blocks. 
    \item \textit{Symmetry} - measure of the combined vertical and horizontal symmetry of a segment/room, computed by comparing pairs of rows and pairs of columns moving outward from the center and summing up the number of positions with the same tile.
\end{itemize}
Both measures were normalized to be between 0 and 1 with 1 representing a fully dense segment and a perfectly symmetrical segment for \textit{Density} and \textit{Symmetry} respectively.

\XFIGUREzelda
\XFIGUREzeldalode
\XFIGUREmetroid
\XFIGUREzeldamet
\XFIGUREzeldamm
\XFIGUREmetmm
\XFIGUREzeldametmm

For each game, we computed these values for each training segment. For evaluation, we sampled a number of latent vectors equal to the number of training segments for that game, and for each vector, computed the mean \textit{Density} and \textit{Symmetry}, averaged across segments generated by conditioning that vector using that game's IN directional labels. For a meaningful comparison between generated and training segments, we opted not to use the OUT labels. For each game, we compared the densities and symmetries of the training segments with those of the generated segments using a Wilcoxon Rank Sum test. Results are shown in Table \ref{XTABLEalldensym}. We observed significant differences (shown in bold) between training and generated segments ($p < .05$) for none of the values for MM, for the 16-dimensional Zelda model in terms of \textit{Density}, for the 16-dimensional LR model in terms of \textit{Density} and the 32-dimensional LR model in terms of both \textit{Density} and \textit{Symmetry} and in terms of all values for Metroid. The absence of statistical significance suggests models being able to successfully capture the structural patterns of the original games, in terms of the measured variables. Thus, the CVAEs for Zelda and MM did the best in terms of modeling the original levels. For LR, the bigger models fared worse than the smaller ones and overall, CVAEs for Metroid did worst in terms of closely capturing the original levels. However, note that the observed mean values for both \textit{Density} and \textit{Symmetry} are close to those of the original values with the standard deviation being far lower for the generated segments. This suggests that Metroid models were able to capture the primary structures and patterns of the original levels but failed in learning an adequate amount of variety and outlier features.

Since most models did well in capturing the structural patterns of the original levels, we wanted to check the performance of the models in terms of generating novel segments. For each game, we sampled 1000 latent vectors and generated segments by conditioning each using all possible directional labels. We then computed what percentage of these segments was not in the original training set, computing separately for IN and OUT labels since we expect more of the IN label-generated segments to be in the training set. Results for this are given in Table \ref{XTABLEnovelty}. The Zelda models produced the least novelty with over half the samples being those from the training set. LR models produced the highest variety, followed by MM, followed by Metroid, producing approximately 80\%, 70\% and 60\% novel segments respectively when using the IN labels. Novelty was higher when using OUT labels but most of these segments likely have incorrect orientations as seen previously. Combining with the prior evaluation, the metrics for Zelda models being closest to training data can be explained by these models generating training segments about half of the time. This can be viewed as good or bad depending on if one prefers generated levels to closely resemble training levels or be more varied. Metroid having the lowest novelty out of the other 3 games but being the farthest from the training data in the prior evaluation reinforces the possibility that the models learned a specific subset of the training data and thus produce those segments when sampled but failed to learn more variations. Thus, future work should look at improving these Metroid models via more game-specific tuning and architectures.

Finally, to evaluate the segments generated by the blended models, we compared the distribution of training segments of the different games with the distribution obtained by using different game conditioning labels. For this comparison, we computed the E-distance between these two sets of segments. E-distance \cite{szekely2013energy} is a measure of similarity between two distributions with lower values indicating higher similarity and has been previously recommended for evaluating generative models \cite{summerville2018expanding}. We used \textit{Density} and \textit{Symmetry} as defined previously for computing E-distance. The idea behind this evaluation is that we expect that, for example, for the Zelda--LR blend, the E-distance between the original Zelda segments and segments generated using the Zelda game label to be lower than the E-distance from segments generated using the LR game label and vice-versa. Results are shown in Figure \ref{XFIGUREedistance}. Except for Metroid-MM, all results are to expectation. When a game is specified in the label, the resulting distribution of generated segments is closer to the training distribution than when it is not. As an example, for Zelda--LR (fig 3.a), E-distance compared to the original LR levels (red line) is lowest when the label is \clabel{0,1} and highest when it is \clabel{1,0}, with similar results for Zelda and the other blends. Results weren't as expected for the Metroid--MM blends however. This may be explained by the \textit{Density} and \textit{Symmetry} values being very similar for the training segments in both games, leading to similar E-distances. Overall, though, we can conclude based on this and the blend label evaluation, that blending in general can be controlled reliably by using these conditioning labels.

\subsectionpre
\subsection{Visual Inspection}
\subsectionpost
Example levels are shown in Figures 4-10. Segments are delineated using white borders with red dashes indicating doors/openings. Non-blended levels (Figures \ref{XFIGUREzelda} and \ref{XFIGUREmetroid}) look a lot more playable while blended levels are less so but arguably more interesting, consisting of smooth blended regions like the Metroid-Zelda sections in Figures \ref{XFIGUREzeldamet} and \ref{XFIGUREzeldametmm} as well as more chaotic regions in Figure \ref{XFIGUREzeldalode}, blending ropes and ladders with doors and dungeons. As-is, this system could be used for generating blended levels for a companion system tasked with generating mechanics that render the levels playable. For e.g., what set of mechanics could one come up with to traverse Zelda dungeons while also being able to execute Metroid jumps and climbing ladders to collect gold? What mechanics make sense in regions that blend Zelda and Metroid? Answering such questions could help move us towards building systems capable of generating entirely new game designs.

\sectionpre
\section{Conclusion and Future Work}

We presented an approach for generating whole platformer levels and dungeons by using CVAEs to control the directionality of generated segments and rooms, in turn also enabling us to take a step towards blending games from different genres, thus opening up several avenues to pursue in the future. 

The main limitation of this work is the lack of playability evaluations. This was confounded by several factors. For Zelda, the VGLC dungeons contain neither lock and key tiles nor the triforce, all of which are crucial for playability. Thus, we restricted ourselves to noting that for Zelda, we obtained near-perfect reliability of door placement, ensuring that generated dungeons are at least traversable from room to room. Regarding Lode Runner, a player completes a level when they've collected all of the gold. However, since we worked with segments, this is harder to check since a segment may be playable but not have gold. Above all, blended levels would warrant new mechanics to be playable and generating mechanics for blended games is a problem we wish to address in the future. The next step is thus to generate fully playable versions of these dungeon-platformer hybrids and blends. We could take levels generated by this work and use a separate process to generate new/blended mechanics to be able to play them. Alternatively, we could use an agent capable of certain mechanics and try repairing a given set of generated levels, using an approach similar to \cite{cooper2020pathfinding}. It would also be fruitful to incorporate this work into a  mixed-initiative or automated design tool where users can supply the labels as well as the proportions for blending. Finally, using different latent sizes did not have much impact both in terms of quantitative results and visual inspection. In the future, we'd like to focus specifically on investigating the effect of different latent sizes.

\vspace{-0.25cm}
\bibliographystyle{IEEEtran}
\bibliography{refs-manual}

\end{document}